\documentclass{article}

\usepackage{microtype}
\usepackage{graphicx}
\usepackage{subcaption}
\usepackage{booktabs} 

\usepackage{hyperref}

\usepackage[preprint]{icml2026}

\usepackage{amsmath}
\usepackage{amssymb}
\usepackage{mathtools}
\usepackage{amsthm}
\usepackage{nccmath}

\usepackage[capitalize,noabbrev]{cleveref}
\usepackage{adjustbox}
\usepackage[table]{xcolor} 
\usepackage{colortbl}      
\usepackage{caption}
\usepackage{multirow}
\usepackage{fancyhdr}

\theoremstyle{plain}

\theoremstyle{definition}

\theoremstyle{remark}

\usepackage[textsize=tiny]{todonotes}

\icmltitlerunning{}

\makeatletter
\def\ps@plain{\ps@empty}
\def\ps@fancy{\ps@empty}
\makeatother

\begin{document}

\twocolumn[
  \icmltitle{SJD-VP: Speculative Jacobi Decoding with Verification Prediction for Autoregressive Image Generation}



  \icmlsetsymbol{equal}{*}

  \begin{icmlauthorlist}
    \icmlauthor{Bingqi Shan}{yyy,pcl}
    \icmlauthor{Baoquan Zhang*}{yyy}
    \icmlauthor{Xiaochen Qi}{comp}
    \icmlauthor{Xutao Li}{yyy}
    \icmlauthor{Yunming Ye}{yyy}
    \icmlauthor{Liqiang Nie}{yyy}
  \end{icmlauthorlist}

  \icmlaffiliation{yyy}{Harbin Institute of Technology, Shenzhen}
  \icmlaffiliation{pcl}{Institute of Perceptual Intelligence, Pengcheng Laboratory, Shenzhen}
  \icmlaffiliation{comp}{SIFAR}
  \icmlcorrespondingauthor{Baoquan Zhang}{baoquanzhang@hit.edu.cn}

  \icmlkeywords{Machine Learning, ICML}

  \vskip 0.3in
]



\printAffiliationsAndNotice{}  

\begin{abstract}
Speculative Jacobi Decoding (SJD) has emerged as a promising method for accelerating autoregressive image generation. Despite its potential, existing SJD approaches often suffer from the low acceptance rate issue of speculative tokens due to token selection ambiguity. Recent works attempt to mitigate this issue primarily from the relaxed token verification perspective but fail to fully exploit the iterative dynamics of decoding. In this paper, we conduct an in-depth analysis and make a novel observation that tokens whose probabilities increase are more likely to match the verification-accepted and correct token. Based on this, we propose a novel Speculative Jacobi Decoding with Verification Prediction (SJD-VP). The key idea is to leverage the change in token probabilities across iterations to guide sampling, favoring tokens whose probabilities increase. This effectively predicts which tokens are likely to pass subsequent verification, boosting the acceptance rate. In particular, our SJD-VP is plug-and-play and can be seamlessly integrated into existing SJD methods. Extensive experiments on standard benchmarks demonstrate that our SJD-VP method consistently accelerates autoregressive decoding while improving image generation quality.
\end{abstract}

\section{Introduction}
\label{sec:intro}


Autoregressive image generation models have achieved remarkable success in producing high-quality images\cite{brown2020language, achiam2023gpt, touvron2023llama}, enabling a wide range of applications such as high-fidelity image synthesis \cite{van2017neural, razavi2019generating, chen2020generative}, controllable image editing \cite{esser2021taming, chang2022maskgit}, text-to-image generation \cite{ramesh2021zero, yu2022scaling}, image inpainting and outpainting \cite{yu2018generative, ramesh2022hierarchical}, and even complex multi-modal reasoning tasks \cite{alayrac2022flamingo, liu2024visual, team2023gemini}. However, their inherently sequential sampling process leads to slow inference\cite{van2017neural, leviathan2023fast}. To address this issue, Speculative Jacobi Decoding (SJD)\cite{SJD} has recently emerged as a promising technique by generating multiple tokens in parallel and validating them through iterative refinement manner. Despite its potential, existing SJD approaches\cite{LANTERN} often deliver limited speedup due to its low speculative acceptance rates, which ultimately constrains their efficiency.

\begin{figure}[t]
    \centering
    \begin{minipage}{0.48\linewidth}
        \centering
        \includegraphics[width=\linewidth, trim=4 4 4 4, clip]{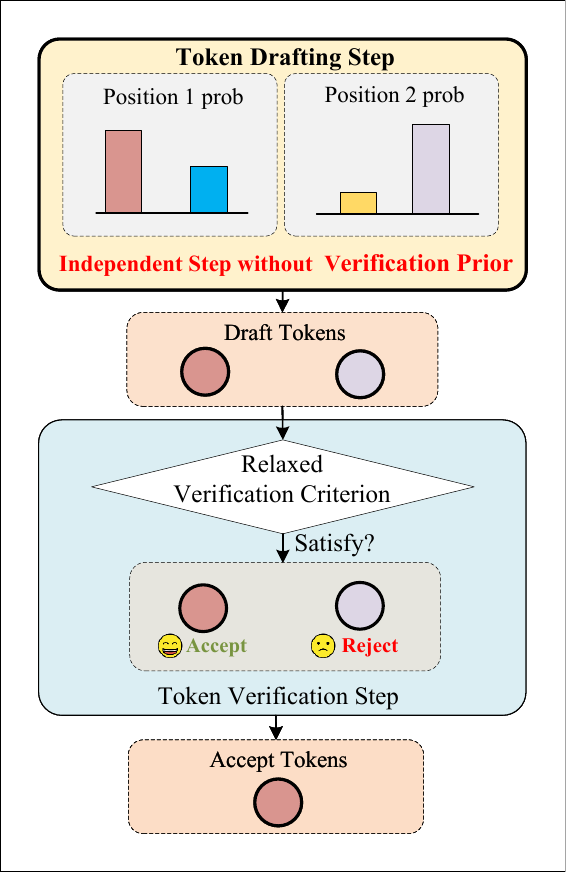}
        \caption*{(a) Existing  method}
        \label{fig1_1}
    \end{minipage}
    \hfill
    \begin{minipage}{0.48\linewidth}
        \centering
        \includegraphics[width=\linewidth, trim=4 4 4 4, clip]{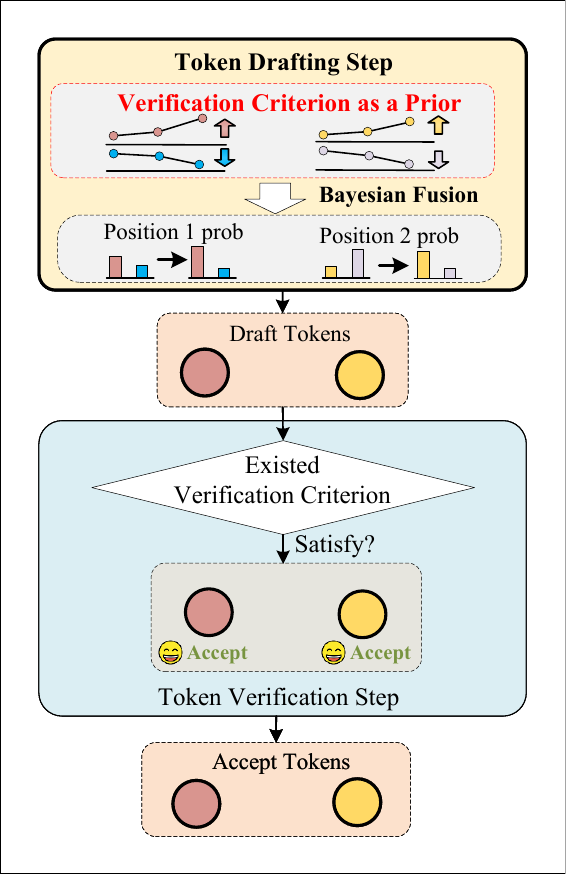}
        \caption*{(b) Our SJD-VP}
         \label{fig1_2}
    \end{minipage}
    \vspace{-2mm}
    \caption{
    Illustration of the structural decoupling in existing SJD and our solution, SJD-VP.
    In (a), the SJD variants drafting step selects tokens based only on the sampling distribution of the current iteration, ignoring the verification history. As a result, the drafting step is decoupled from the verification step, often leading to rejected tokens.
    In contrast, (b), our SJD-VP actively leverages verification criterion as a prior by analyzing probability trajectories of tokens. By predicting verification preference and guiding the distribution to favor consecutive probability growth tokens, it achieves alignment between the drafting and verification steps.
    }
    \label{fig:intro}
    \vspace{-2mm}
\end{figure}

Recently, existing studies point out that the main reason causing such low acceptance rates is token semantic ambiguity issue, i.e., image tokens often exhibit inherent redundancy and diversity, meaning multiple tokens can convey similar semantics. Motivated by this observation, these existing SJD variants such as LANTERN\cite{jang2024lantern} and GSD\cite{so2025grouped} attempt to improve the acceptance rate during speculative Jacobi decoding by relaxing the verification criterion to tolerate semantically similar tokens. Although these methods have shown superior performance, as shown in Figure~\ref{fig:intro}(a), they still suffer from another key limitation: the drafting step and the verification step are usually treated as two independent processes. This structural decoupling leads to an inherent misalignment between the drafted tokens and the verification criterion, making it difficult to generate the draft tokens satisfying the deterministic verification criterion. In other words, the drafting step does not take into account whether the generated tokens are likely to be accepted by the verification step, resulting in a low hit rate of valid tokens and thus limiting the overall acceleration potential. This naturally raises a key question: \emph{Can the deterministic verification criterion be leveraged to enable Verification Prediction, i.e., guiding the sampling process to generate draft tokens that are more likely to pass verification and thereby further improving efficiency?}

To answer this question, in this paper, we conduct an in-depth analysis (see Section~\ref{sec:motivation-analysis} for more details) on these verification-accepted tokens, i.e., those token satisfying the deterministic verification criteria. Then, we reveal a key phenomenon, i.e.,
\textbf{most verification-accepted tokens show a consecutive probability growth along the Jacobi iterations before performing the verification step and such phenomenon can be consistently observed in nearly 95\% of all verification-accepted tokens.} This phenomenon means that such consecutive probability growth pattern can be effectively leveraged to enable the verification prediction, i.e., guiding the generation of draft tokens which are more likely to pass verification and thus improve the overall acceptance rate.
However, a natural follow-up question arises: Are the tokens predicted based on this growth pattern actually the correct tokens? To figure out this point, we conducted further experiments on these correct tokens and observe that \textbf{this consecutive probability growth is highly specific to correct tokens, whereas incorrect tokens typically exhibit fluctuating or decreasing probabilities.} This indicates that tokens following such a consecutive probability growth are also more likely to be the correct tokens.

The above two-point analysis provides a key insight: \emph{by drafting tokens along the probability growth observed during Jacobi iterations, we can not only increase their likelihood of passing subsequent verification but also guide them toward the correct tokens, thereby simultaneously accelerating decoding and improving the quality of the generated images.}  Based on this insight, we propose a novel speculative Jacobi decoding with verification prediction (called SJD-VP) for accelerating autoregressive image generation. As illustrated in Figure~\ref{fig:intro}(b), our core idea is to regarding the verification prior estimated from token probability growth patterns as a Bayesian prior and viewing the current drafting probability as the conditional likelihood, and then leverage the Bayesian manner to derive a posterior as the final sampling probability, thereby achieving faster decoding while simultaneously improving the generation quality. In particular, our SJD-VP is very simple and plug-and-play, which can be seamlessly integrated into existing SJD and its variants by only conducting few modifications in token drafting step. Specifically, we first execute \textit{Verification Prior Estimation} to estimate a verification prior from historical probability trajectories, identifying tokens that show consecutive probability growth. Then, we perform \textit{Bayesian Fusion} to derive a posterior sampling distribution by fusing this verification prior with the current drafting distribution. Finally, we sample from this posterior distribution, thereby effectively ensuring that the drafted step not only align with the verification step but are also significantly higher quality.


Our main contributions can be summarized as follows:
\begin{itemize}
    \item  We identify that overlooking the prior of the verification criterion is a key factor constraining the efficiency of SJD and its variants. We reveal that this prior can be leveraged to predict which tokens are more likely to be accepted during the drafting step, offering a new perspective to further enhance decoding efficiency.

    \item We propose \textbf{SJD-VP}, a plug-and-play sampling mechanism that integrates \textit{verification prediction} into the decoding step via a Bayesian framework. By fusing the verification prior with the current drafting probability to predict verification results, it improves the acceptance rate of draft tokens. Importantly, our method is complementary to methods that relax the verification criterion, and can be integrated with them.

    \item Extensive experiments demonstrate that SJD-VP consistently accelerates decoding while improving image quality. Furthermore, it significantly boosts the performance of SJD variants, thereby robustly validating effectiveness and general flexibility of SJD-VP.
\end{itemize}

\section{Related Work}
\label{sec:relatedwork}
\subsection{Autoregressive Image Generation}
Autoregressive image generation treats an image as a sequence of visual tokens and predicts each token in an autoregressive manner. This sequential formulation captures fine-grained dependencies and has become a core paradigm for high-fidelity image synthesis. Existing autoregressive image generation methods can be roughly divided into two groups: \textbf{1) Continuous-token autoregressive models.} These approaches operate in a continuous embedding space and directly autoregress over pixel values or latent feature vectors, avoiding quantization into discrete codebooks, such as PixelRNN\cite{van2016pixel}, PixelCNN\cite{van2016conditional}, iGPT\cite{chen2020generative}, DALL·E 2 prior \cite{ramesh2022hierarchical}. \textbf{2) Discrete-token autoregressive models.} This line of works aim to quantize images into discrete token sequence and then autoregressively generate them in this discrete token space. Representative methods include VQ-VAE, VQ-GAN \cite{van2017neural, esser2021taming}, DALL·E \cite{ramesh2021zero}, and Parti \cite{yu2022scaling}, whose learned discrete codebooks enable Transformer to synthesize high-fidelity images. Recent multimodal framework further extend this paradigm by unifying text and image tokens into a shared vocabulary for cross-modal generation, such as Janus and Lumina-mGPT \cite{liu2024lumina, wu2025janus}. Although these existing methods have achieved strong controllability and fidelity, they remain constrained by the inherently slow sequential decoding process. Consequently, addressing this inference latency has become a critical research priority. In this paper, we also focus on accelerating this decoding process.

\subsection{Autoregressive Generation Acceleration}
Autoregressive generation acceleration aims to reduce the latency caused by step-by-step token decoding, which becomes especially prohibitive for high-resolution image synthesis. Existing acceleration approaches can be roughly divided into two groups: \textbf{1) Model-based acceleration methods.} These approaches improve autoregressive image generation efficiency by redesigning model architectures or introducing auxiliary predictors during training . For example, in \cite{wang2025parallelized, ren2025beyond, yu2024magvit2}, they explore hierarchical token prediction, distillation from large AR transformers, and hybrid AR–parallel generation modules, respectively, to enable faster autoregressive generation inference while maintaining high generation fidelity;
\textbf{2) Inference-based acceleration methods.}
These approaches speed up autoregressive generation purely at inference time without modifying or retraining the model. A representative line of work is speculative decoding~\cite{leviathan2023fast}, where draft tokens are proposed in parallel and verified by the base model. Speculative Jacobi Decoding (SJD)~\cite{SJD} extends this paradigm to discrete autoregressive image generation, using parallel speculative updates to significantly reduce decoding steps. However, SJD suffers from low acceptance rates which limits its acceleration. Recent studies like LANTERN~\cite{jang2024lantern} and GSD~\cite{so2025grouped} propose to relax the verification criterion to accept semantically similar tokens. These variants improve the acceptance rate of SJD while maintaining image quality. 

Nevertheless, as discussed in Section ~\ref{sec:intro}, these methods still suffer from structural decoupling between the drafting step and the verification step. In this paper, we propose SJD-VP to address this issue by introducing \textbf{verification prediction} into the drafting step. Since our method does not modify the verification criterion, it \textbf{complements existing verification methods} to yield substantial further acceleration gains.

\section{Preliminaries}
\subsection{Problem Definition}
We aim to accelerate autoregressive image generation while preserving fidelity. Formally, given a conditioning signal $x$, a decoder models the token sequence $y = (y_1, \dots, y_T)$ via:
\begin{equation}
p(y \mid x) = \prod_{t=1}^{T} p(y_t \mid y_{<t}, x).
\label{eq:ar_factorization}
\end{equation}
Standard inference incurs $O(T)$ sequential steps, creating a significant latency bottleneck for high-resolution images. In this work, we focus on the \emph{inference-only} setting, where the pretrained model remains fixed without any additional retraining or architectural modifications. Within this context, Speculative Jacobi Decoding (SJD) has emerged as a leading paradigm. Next, we briefly review existing SJD below.


\subsection{Speculative Jacobi Decoding}
Speculative Jacobi Decoding (SJD)~\cite{SJD} is a training-free acceleration strategy for efficient autoregressive image generation. Specifically, SJD follows a "draft-then-verify" paradigm, i.e., each iteration first drafts a window of $W$ future tokens in conditioned on the prefix $c$, and then sequentially verifies them against the base model. Valid tokens are accepted to update the context, while the first rejection terminates the window and triggers resampling. This cycle repeats until generation completes. 

Although SJD has shown superior performance, SJD suffers from a critical limitation: the drafting step is decoupled from the verification step, such that the drafting step does not take into account whether the generated tokens are likely to be accepted by the verification step. This may be a key reason causing a low acceptance rate of valid tokens and thus limiting the overall acceleration potential. Therefore, a natural question is: \emph{Can the deterministic verification criterion be leveraged to enable Verification Prediction, i.e., guiding the token drafting step to generate tokens that are more likely to pass verification and thereby further improving efficiency?}

\section{Methodology}
\subsection{Motivation Analysis}
\label{sec:motivation-analysis}

\begin{figure}[t]
    \centering
    \begin{subfigure}[t]{0.55\linewidth}
        \centering
        \includegraphics[width=\linewidth, trim=5mm 5mm 5mm 5mm, clip]{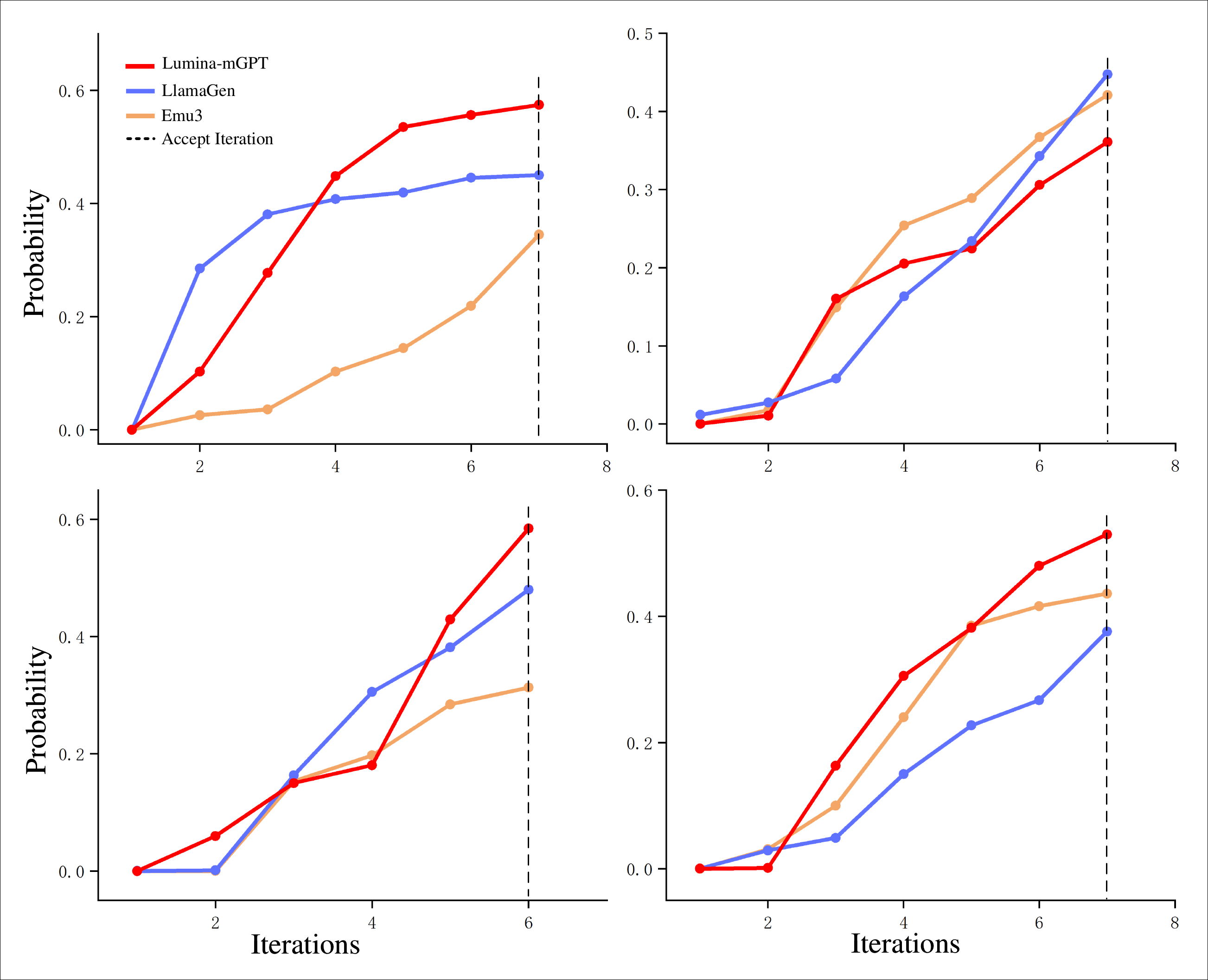}
        \caption{Tokens' probability trajectories}
        \label{fig:sub_a}
    \end{subfigure}
    \hfill
    \begin{subfigure}[t]{0.43\linewidth}
        \centering
        \includegraphics[width=\linewidth, trim=5mm 5mm 5mm 5mm, clip]{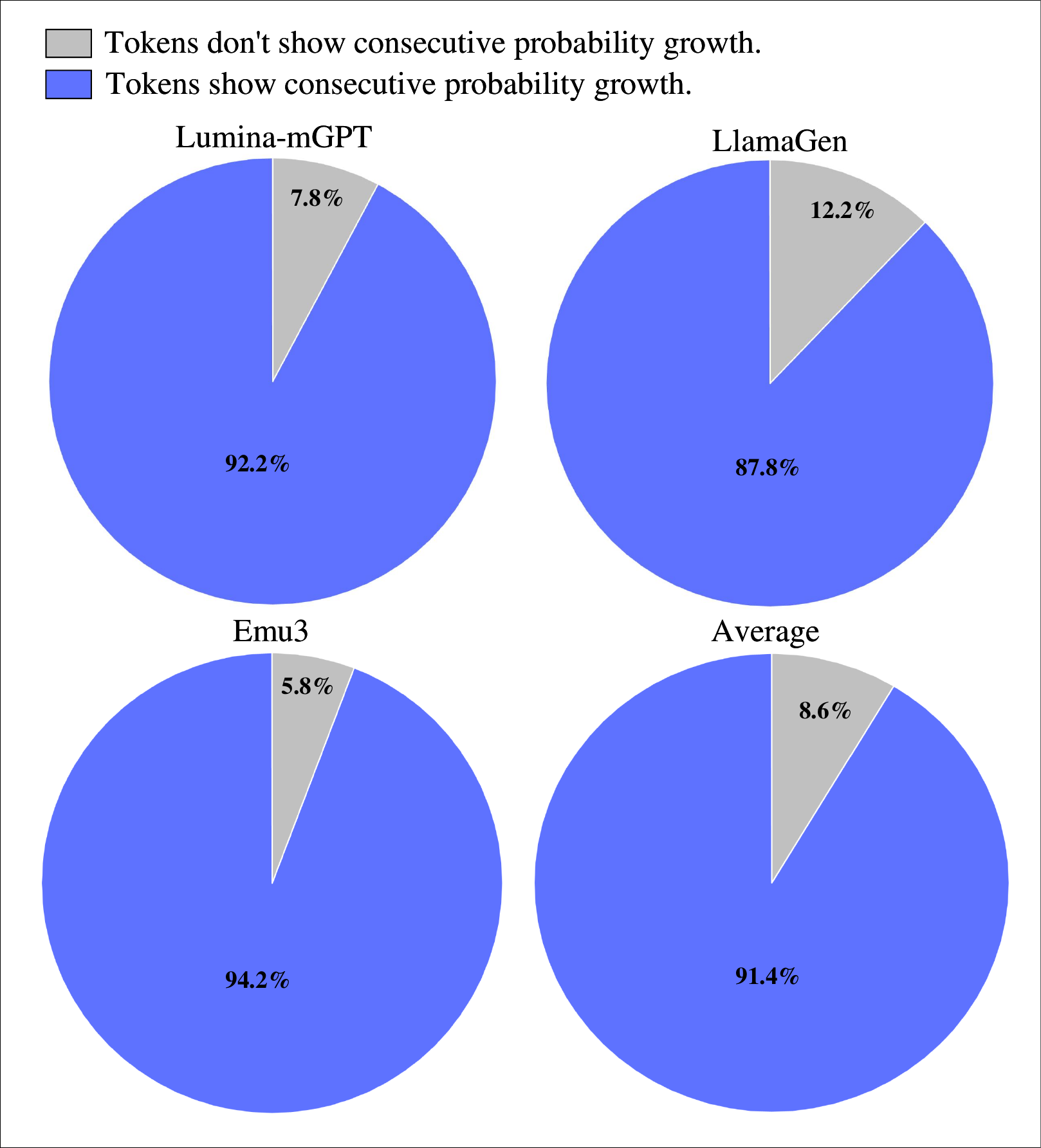}
        
        \caption{Statistical analysis}
        \label{fig:sub_b}
    \end{subfigure}
    \vspace{-2mm}
    \caption{Motivation Analysis on Verification-Accepted Tokens. 
    }
    \label{fig:all_visual}
    \vspace{-5pt}
\end{figure}



\paragraph{Analysis on Verification-Accepted Tokens.} To answer the above question, in this section, we conduct an in-depth analysis on these verification-accepted tokens (i.e., these tokens satisfying the deterministic verification criteria). Specifically, we first take the latest AR image models (i.e., Lumina-mGPT~\cite{liu2024lumina-mgpt}, LlamaGen~\cite{sun2024autoregressive}, and Emu3~\cite{wang2024emu3}) with Speculative Jacobi Decoding as the base model. Then, we randomly select four text prompts from PartiPrompts dataset and visualize its probability trajectory of verification-accepted tokens across multiple Jacobi iterations, respectively. The results are shown in Figure~\ref{fig:all_visual}(a). From results, we can see that these verification-accepted tokens usually present a consecutive probability growth before it is verification-accepted. This means the character ``consecutive probability growth" in iterative dynamics of decoding can be regarded as a prediction prior for these verification-accepted tokens. 

To further assess whether this character ``consecutive probability growth" can be regarded as a reliable prediction prior for the verification step, in Figure~\ref{fig:all_visual}(b), we further present a statistical analysis of the proportion of verification-accepted tokens that demonstrate consecutive probability growth before being accepted. From results, we can see that From the results, we observe that in Lumina-mGPT, LlamaGen, and Emu3, approximately 92.3\%, 87.7\%, and 94.2\% (with an average of 91.4\%) of verification-accepted tokens demonstrate this consecutive probability growth pattern. Such consistently high proportion across different models indicates that the consecutive probability growth is a strong and model-agnostic character associated with successful verification. This suggest that \emph{this ``consecutive probability growth"  can serve as an effective prediction prior for the verification step, providing a principled basis for guiding the drafting process toward high-acceptance tokens and thereby improving the overall efficiency of speculative decoding.}

\begin{figure}[t]
    \centering

    \begin{subfigure}{0.48\linewidth}
        \centering
        \includegraphics[width=\linewidth]{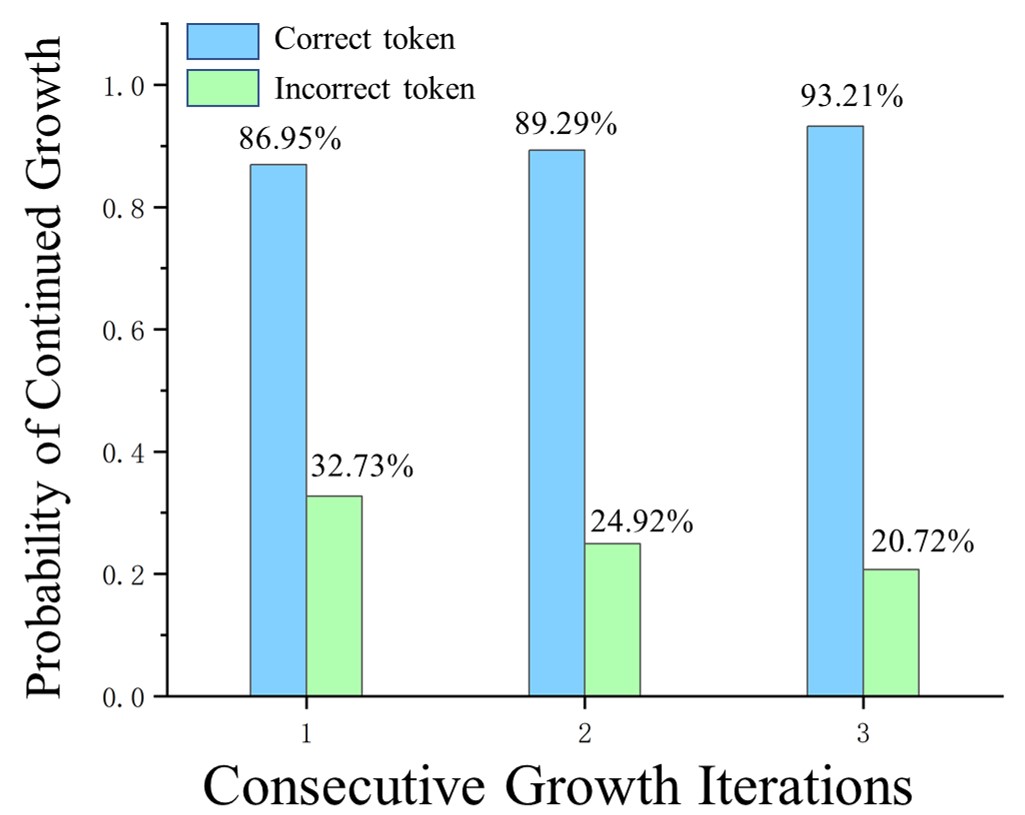}
        \caption{Proportion of correct and incorrect tokens with consecutive $n$-steps probability growth.}
        \label{fig:trend_cont_a}
    \end{subfigure}
    \hfill
    \begin{subfigure}{0.48\linewidth}
        \centering
        \includegraphics[width=\linewidth]{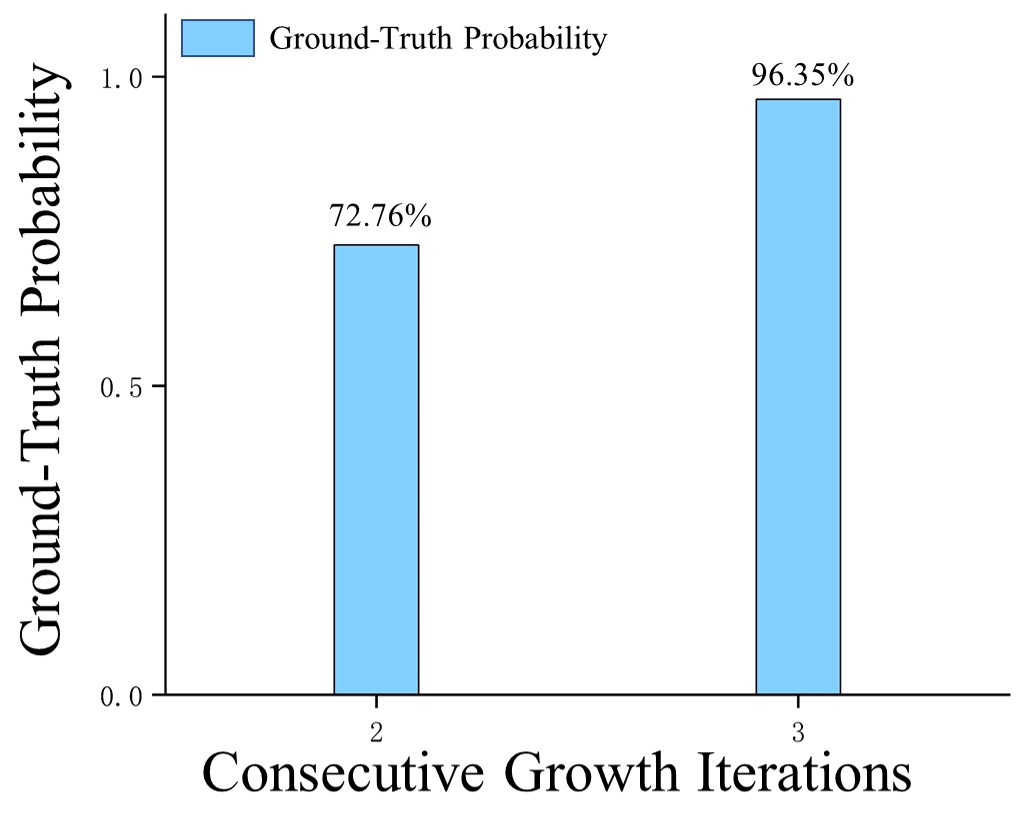}
        \caption{Prediction accuracy of correct tokens with ``consecutive $n$-steps probability growth".}
        \label{fig:trend_cont_b}
    \end{subfigure}

    \caption{ Motivation Analysis on Correct Tokens.
    }
    \label{fig:additional_prediction}
    \vspace{-20pt}
\end{figure}

\paragraph{Analysis on Correct Tokens.} However, a natural follow-up question is: \emph{Are the tokens predicted based on the above character ``consecutive probability growth" actually the correct tokens? }
To figure out this point, in Figure~\ref{fig:additional_prediction}, we
further conduct experiments on these correct and incorrect tokens.
Specifically, as shown in Figure~\ref{fig:additional_prediction}(a),  given that a token has exhibited probability increases for
$n$ consecutive steps, we compute the probability that its likelihood will continue to increase at the next step. From results, we can see that 1) these correct tokens (see blue bars) consistently demonstrate a much higher continuation probability exceeding 85\% for all values of $n$; whereas 2) the continuation probability of incorrect tokens (see green bars) rapidly decreases as $n$ grows. These results quantitatively confirm that the consecutive probability growth can be regarded as a distinctive and reliable characteristic of correct tokens. 

Finally, to further evaluate whether this prediction prior ``consecutive probability growth" can directly guide correct token drafting, as shown in Figure~\ref{fig:additional_prediction}(b), we measure the probability that a token selected solely based on having consecutive probability growth for $n$ iterations corresponds to the correct token. As shown in Figure~\ref{fig:additional_prediction}(b), this probability is already relatively high (72.76\%) when consecutive growth first emerges ($n=2$), and increases substantially to 96.35\% as the growth persists ($n=3$). This means that \emph{token drafting based on prior of consecutive probability growth not only enables predict verification results to improve efficiency but also guides sampling toward correct tokens to preserve image generation quality.}

\subsection{The Proposed SJD-VP Method}

As discussed in Section~\ref{sec:intro}, existing methods usually regard drafting and verification steps as two independent processes, which leads to a misalignment issue between drafted tokens and verification criterion. To this end, we propose a novel speculative Jacobi decoding with verification prediction (SJD-VP), which regards the verification prior (i.e., the character ``consecutive probability growth") estimated from verification criterion as a Bayesian prior and the current drafting probability as the conditional likelihood, and then leverage the Bayesian manner to derive a posterior as final sampling probability, thereby accelerating autoregressive generation. In particular, our method only revise the token drafting step, which is plug-and-play and can be seamlessly integrated into existing SJD methods. Next, we take the existing SJD as example to introduction our SJD-VP and how to adapt to other SJD variants will be explained at the end of the section. Specifically, as shown in Algorithm~\ref{alg:sjdvp}, for each token drafting step, we first perform the \textbf{Verification Prior Estimation} phase to construct a historical reference $\bar{p}_t(x)$ and leverage it to estimate the verification prior $S_t(x)$ with the character ``consecutive probability growth". Then, we perform the \textbf{Bayesian Verification Fusion} phase to fuse the prior with the current drafting probability to derive a posterior probability $p't(x)$. Finally, we sample the token from the posterior probability $p't(x)$ to draft tokens. Next, we detail on these two phases, respectively.
\paragraph{Verification Prior Estimation} As discussed in Section~\ref{sec:motivation-analysis}, verification-accepted tokens exhibit a pattern of consecutive probability growth. Therefore, we can estimate verification prior by capturing such growth trend between the current probability and its historical probability. Specifically, let $p_t(x)$ denote the probability of a candidate token $x$ at the $t$-th Jacobi iteration. We maintain a position-aware history buffer of length $L$ to record the recent probability trajectories $\{p_{t-L}(x), \dots, p_{t-1}(x)\}$. Then, we employ an Exponentially Weighted Aggregation (EWA) to synthesize a robust historical reference probability $\bar{p}_t(x)$ that aggregates historical Jacobi-iteration probability to mitigate volatile fluctuations between iterations while remaining sensitive to recent Jacobi-iteration probability. That is,

\begin{equation}
    \bar{p}_t(x) = \frac{\sum_{k=0}^{L} \gamma^k \cdot p_{t-k}(x)}{\sum_{k=0}^{L} \gamma^k},
\end{equation}
where $\gamma \in (0, 1)$ is the decay factor of EWA. 

With the above historical probability $\bar{p}_t(x)$, we first calculate a prediction score $S_t(x)$ by computing the gain between current probability $p_t(x)$ and historical probability $\bar{p}_t(x)$.  Considering the mathematical stability, we perform this gain computation in the log-space. That is,

\begin{equation}
    S_t(x) = \log p_t(x) - \log \bar{p}_t(x).
\end{equation}

\begin{algorithm}[tb]
   \caption{The Proposed SJD-VP Method}
   \label{alg:sjdvp}
\begin{algorithmic}
   \footnotesize 
   \STATE {\bfseries Input:} Model $\theta$, Prefix $c$, Window $W$, History buffer $\mathcal{H}$, Buffer length $L$, Decay $\gamma$, Top-$K$ ratio, Current step $t$, Current probabilities $p_t(x)$
   \STATE {\bfseries Output:} Sequence $X$
   
   \WHILE{sequence $X$ not finished}
       \STATE{\textbf{ --- Token Drafting Step ---}}
       \STATE  \textit{// \textbf{Phase 1}: Verification Prior Estimation }
       \STATE Record historical trajectories $\{p_{t-L}, \dots, p_{t-1}\}$ from $\mathcal{H}$;
       
       \STATE Calculate the historical probability $\bar{p}_t(x)$ by Eq. (2);
    
       \STATE Estimate the verification prior $S_t(x) \leftarrow M_t(x)(\log p_t(x) - \log \bar{p}_t(x))$ with consecutive growth trend by Eqs. (3) and (4);
       
       
       \STATE  \textit{// \textbf{Phase 2}: Bayesian Verification Fusion}
       \FOR{each candidate $x$ in Vocabulary $\mathcal{V}$}
           \IF{$x \in$ Top-$k$ }
               \STATE $\log p'_t(x) \leftarrow \log p_t(x) + M_t(x) \cdot S_t(x)$ \quad// Eq. (5)
           \ELSE
               \STATE $\log p'_t(x) \leftarrow \log p_t(x)$ \quad// Eq. (5)
           \ENDIF
       \ENDFOR
       
       \STATE \textit{Perform token drafting from $p'_t(x)$ by Eq. (6)}
       \STATE \textbf{--- Token Verification Step ---}
       \STATE Verify drafted tokens against $\theta$ (Same as Algorithm~\ref{alg:sjd})
       \STATE Update prefix $c \leftarrow c + \text{accepted tokens}$
       \STATE Update History $\mathcal{H}$ with current distributions
   \ENDWHILE
   
   \STATE $X \leftarrow c$
   \STATE \textbf{return} $X$
\end{algorithmic}
\end{algorithm}

However, the verification prior estimated by this comparison does not explicitly validate whether a token shows consecutive probability growth. To address this issue, we further introduce a \textbf{Consecutive Growth Mask $M_t(x)$} to mark whether a token $x$ shows consecutive probability growth:
\begin{equation}
    M_t(x) = 
    \begin{cases} 
    1, & \text{if } p_{t-k}(x) > p_{t-k-1}(x), \\
    0, & \text{otherwise},
    \end{cases}
\end{equation}
where the condition must hold for all $k \in \{0, \dots, N\}$. By applying this mask, we obtain the \textbf{Verification Prior} $M_t(x) \cdot S_t(x)$. It filter out tokens that lacking the required growth and is used for Bayesian Verification Fusion.

\paragraph{Bayesian Verification Fusion} Until, we have obtained two probability, i.e., the probability $p'_t(x)$ estimated by verification prior and the current observed probability $p_t(x)$. To achieve more accurate sampling probability, we regard the the prior probability $p'_t(x)$ as a Bayesian prior and the current probability $p_t(x)$ as a conditional likelihood, and then estimate the Bayesian posterior probability $p'_t(x)$ as our final sampling probability. In particular, we perform such Bayesian fusion only on the Top-$k$ candidates (denoted as $\mathcal{C}_t$) for robustness. Similarly, we perform such operation in the log space for mathematical stability. That is,

\begin{equation}
\label{eq:sjd-vp}
    \log p'_t(x) = 
    \begin{cases} 
    \log p_t(x) + M_t(x) \cdot S_t(x), &\text{if } x \in \mathcal{C}_t, \\
    \log p_t(x), & \text{otherwise}.
    \end{cases}
\end{equation}
Finally, we sample the next token $x_t$ from the posterior probability $p'_t(x)$ via a standard Softmax. That is,
\begin{equation}
    x_t \sim \text{Softmax}(\log p'_t(x)).
\end{equation}

\subsection{Adaption to Other SJD Variants}
Our method is plug-and-play, which can be directly integrated into other SJD variants(e.g., LANTERN and GSD), by directly replacing their original token draft step with our token draft step with verification prediction.

\subsection{Theoretical Proof for Effectiveness of SJD-VP}
The goal of our SJD-VP is to improve the acceptance rate of draft tokens during verification step. 
Next, we explain why SJD-VP can achieve this goal from a theoretical perspective. 
Formally, following \cite{so2025grouped}, improving the acceptance rate of draft tokens is theoretically equivalent to reducing the Total Variation Distance ($D_{TV}$) between draft distribution $p_t$ and verification distribution $q$, i.e.,
\begin{equation}
    D_{\text{TV}}(p_t, q) = \frac{1}{2} \sum_{x} |p_t(x) - q(x)|.
\end{equation} 
Therefore, demonstrating the effectiveness of SJD-VP is equivalent to proving that it can reduce the Total Variation Distance, i.e., proving $D_{TV}(p'_t, q) < D_{TV}(p_t, q)$, where $p'_t$ is the posterior probability estimation by our SJD-VP.

\paragraph{The Derivation of $D_{TV}(p'_t, q)=\frac{1}{2} \sum_{x} |p'_t(x) - q(x)|$.} In particular, the posterior estimation $p'_t$ can be understood as a perturbation of the original draft distribution $p_t(x)$, i.e.,
\begin{equation}
    p'_t(x) = p_t(x) + \Delta p(x),
    \label{eq9}
\end{equation}
where $\Delta p(x)$ represents a probability perturbation applied to each token from the prior of our SJD-VP to better align $p_t$ with the validation distribution $q$. Formally, the $\Delta p(x)$ obtained by our SJD-VP can be regarded as first predicting an adjustment direction $\hat{y}(x) \in \{-1, +1\}$ for each token, indicating whether its probability should be increased ($+1$) or decreased ($-1$). The magnitude of the shift is controlled by a small parameter $m>0$ and optionally weighted by $\omega(x)\ge 0$ to reflect token importance. Therefore, the small probability shift $\Delta p(x)$ can be expressed as:
\begin{equation}
    \Delta p(x) = m \cdot \omega(x) \hat{y}(x).
    \label{eq10}
\end{equation}
Therefore, according to Eqs. \ref{eq9} and \ref{eq10}, the Total Variation Distance of our SJD-VP can be further expressed as:
\begin{equation}
\small
\begin{split}
    D_{\text{TV}}(p'_t, q) &= \frac{1}{2} \sum_x |p'_t(x) - q(x)| \\
    &= \frac{1}{2} \sum_x |p_t(x) + m \cdot \omega(x) \hat{y}(x) - q(x)|.
\end{split}
\end{equation}
Then, we can perform a first-order Taylor expansion at $m=0$ and further simplify the $D_{\text{TV}}(p'_t, q)$ as:

\begin{equation}
\label{taylor}
\medmath{
\begin{split}
    &D_{\text{TV}}(p'_t, q) \approx D_{\text{TV}}(p_t, q) + m \frac{d}{dm} D_{\text{TV}}(p'_t, q) \Big|_{m=0} \\
    &\approx D_{\text{TV}}(p_t, q) + m \frac{1}{2} \sum_x \text{sign}(p_t(x)-q(x)) \cdot \omega(x) \hat{y}(x). 
\end{split}
}
\end{equation}

\paragraph{Theoretical Proof on $D_{TV}(p'_t, q) < D_{TV}(p_t, q)$.} To formalize the correctness of the predicted adjustment, we define the ideal adjustment direction for each token as
\begin{equation}
    y(x) = \text{sign}(q(x) - p_t(x)),
\end{equation}
where $y(x)=+1$ indicates an underestimated token and $y(x)=-1$ indicates an overestimated token.  
The prediction accuracy $Q$ of SJD-VP is measured by how often the predicted direction $\hat{y}(x)$ matches the ideal direction $y(x)$:
\begin{equation}
    Q = \mathbb{P}_{x \sim p_t}[\hat{y}(x) = y(x)].
\end{equation}

Substituting the identity $-y(x) = \text{sign}(p_t(x) - q(x))$ into the Taylor expansion in Eq.~\ref{taylor}, we explicitly establish the relationship of $D_{\text{TV}}(p'_t, q)$ with the ideal direction $y(x)$:
\begin{equation}
\label{eq14}
\medmath{
\begin{split}
    &D_{\text{TV}}(p'_t, q) \approx D_{\text{TV}}(p_t, q) - m \frac{1}{2} \sum_x y(x) \cdot \omega(x) \hat{y}(x).  \\
    &\text{Taking expectation over } x \sim p_t: \\
    &\approx D_{\text{TV}}(p_t, q) - \frac{m}{2} \mathbb{E}_{x \sim p_t} [y(x)\hat{y}(x) \omega(x)] \\
    &\approx D_{\text{TV}}(p_t, q) - \frac{m}{2} (\mathbb{E}[y(x)\hat{y}(x) ]\mathbb{E}[\omega(x)]+Cov(y\hat{y},\omega))
\end{split}
}
\end{equation}

Given that the product term $y(x)\hat{y}(x)$ equals $+1$ when the prediction is correct and $-1$ is incorrect, we have:
\begin{equation}
\label{eq15}
\mathbb{E}_{x \sim p_t}[y(x)\hat{y}(x)] = 1 \cdot Q + (-1) \cdot (1 - Q) = 2Q - 1.
\end{equation}
Substituting the expectation derived in Eq.~\ref{eq15} back into the Taylor expansion in Eq.~\ref{eq14}, and considering that $\omega(x)$ is a strictly positive weighting term. we obtain the final evolution of the Total Variation distance:
\begin{equation}
\label{eq16}
\small
    D_{\text{TV}}(p'_t, q) \approx D_{\text{TV}}(p_t, q) - \frac{m}{2} ((2Q - 1) \mathbb{E}[\omega(x)] + Cov(y\hat{y},\omega)).
\end{equation}
Specifically, since the non-negative $\omega(x)$ is determined solely by the probability growth pattern (Eq.\ref{eq:sjd-vp}), it is statistically independent of the prediction results $y\hat{y}$. This independence implies that the $Cov(y\hat{y},\omega) \approx 0$. Hence, the validity of $D_{TV}(p'_t, q) < D_{TV}(p_t, q)$ is governed by the term $(2Q-1)$. Empirical analysis (see Figure~\ref{fig:additional_prediction}(b)) confirms that the prediction accuracy $Q$ based on consecutive growth is at least $72.76\%$, significantly higher than $0.5$, ensuring $(2Q-1) > 0$. This validates the effectiveness of SJD-VP.

\section{Experiments}
\begin{table*}[t!]
    \centering
    \caption{Results of quantitative evaluation on the Parti-prompt and MS-COCO 2017 benchmarks.. The best results are highlighted in bold. }
    \setlength{\tabcolsep}{17pt}
    \resizebox{0.95\linewidth}{!}{ 
    \begin{tabular}{l c c c c c c}
        \toprule
        \textbf{Configuration} & \textbf{Latency (↓)} & \textbf{NFE (↓)} & \multicolumn{2}{c}{\textbf{Acceleration} (↑)} & \textbf{FID (↓)} & \textbf{CLIP-Score (↑)} \\
        \cmidrule(lr){4-5}
        & & & \textbf{Latency} & \textbf{NFE} & & \\
        \midrule
        \multicolumn{7}{l}{\textbf{Parti-prompt}} \\
        Lumina-mGPT \cite{luminangpt} & 79.37s & 2392 & 1.00× & 1.00× & -- & 32.09  \\
        Jacobi Decoding \cite{jacobi} & 82.21s & 2300.0 & 0.97x & 1.04× & -- & 32.09 \\
        \midrule
        SJD \cite{SJD} & 36.07s & 1035.3 & 2.20x & 2.31× & -- & 32.09 \\
        \rowcolor{gray!20}
        SJD-VP & \textbf{27.00s} & \textbf{720.65} & \textbf{2.94x} & \textbf{3.32x} & -- & \textbf{32.15} \\
        \midrule
        LANTERN\cite{jang2024lantern} & 31.40s & \textbf 636.66 & 2.53x & 3.76x & -- & 32.10 \\
        \rowcolor{gray!20}
        LANTERN-VP & \textbf{25.97s} & \textbf{580.47} & \textbf{3.06x} & \textbf{4.12x} & -- & \textbf{32.11} \\
        \midrule
        GSD \cite{so2025grouped} & 33.36s & \textbf 898.97 & 2.38x & 2.66x & -- & 32.11 \\
        \rowcolor{gray!20}
        GSD-VP & \textbf{31.10s} & \textbf{671.60} & \textbf{2.55x} & \textbf{3.56x} & -- & \textbf{32.13} \\
        \rowcolor{gray!20}

        \bottomrule
        \midrule
        \multicolumn{7}{l}{\textbf{MS-COCO 2017}} \\
        Lumina-mGPT \cite{luminangpt} & 86.55s & 2379 & 1.00× & 1.00× & 30.79 & 31.31  \\
        Jacobi Decoding \cite{jacobi} & 85.64s & 2312 & 1.01x & 1.03x & 30.78 & 31.31 \\
        MC-SJD \cite{so2025mc} & 33.28s & 814.5 & 2.60x & 2.92x & 30.73 & 31.32 \\
        \midrule
        SJD \cite{SJD} & 40.10s & 1058.6 & 2.16x & 2.25× & 30.78 & 31.31 \\
        \rowcolor{gray!20}
        SJD-VP &  \textbf{26.95s} & \textbf{729.70}  & \textbf{3.21x} & \textbf{3.26x} & \textbf{30.77} & \textbf{31.42} \\ 
        \midrule
        LANTERN\cite{jang2024lantern} & 33.80s &  655.37 & 2.56x & 3.63x & 31.72 & 31.32 \\
        \rowcolor{gray!20}
        LANTERN-VP &  \textbf{26.71s} & \textbf{590.33}  & \textbf{3.24x} & \textbf{4.03x} & \textbf{31.70} & \textbf{31.43} \\ 
        \midrule
        GSD \cite{so2025grouped} & 34.12s & 925.89 & 2.54x & 2.57x & 31.50 & 31.33 \\
        
        \rowcolor{gray!20}
        GSD-VP & \textbf{30.78s}  & \textbf{689.58} & \textbf{2.81x} & \textbf{3.45x} & \textbf{31.43} &  \textbf{31.46} \\
        \rowcolor{gray!20}
        \bottomrule
        \midrule
    \end{tabular}
    }
    \label{tab:main}
    \vspace{-0.3cm}
\end{table*}

\subsection{Experimental Settings}
\paragraph{Dataset}
We evaluate SJD-VP on two standard benchmarks: (1) \textbf{Parti-prompts}~\cite{yu2022scaling}, containing 1,600 diverse open-domain prompts; and (2) \textbf{MS-COCO 2017}~\cite{mscoco}, comprising 118K image-caption pairs. For both, we strictly follow the preprocessing and evaluation protocols established in prior works~\cite{SJD,so2025grouped}. All experiments use identical sampling conditions to ensure fair comparison. 
\paragraph{Baselines and Metrics.}
We compare SJD-VP against the base autoregressive model \textbf{Lumina-mGPT}~\cite{luminangpt} and four state-of-the-art decoding strategies: \textbf{Jacobi Decoding}~\cite{jacobi}, \textbf{SJD}~\cite{SJD}, \textbf{LANTERN}~\cite{jang2024lantern}, and \textbf{GSD}~\cite{so2025grouped}. All methods utilize the same backbone and hyperparameters to ensure fair comparison. 
Performance is evaluated by using five standard evaluation metrics. \textbf{Latency} (in seconds per image) and \textbf{NFE} (number of function evaluations) quantify decoding efficiency, 
while \textbf{Acceleration} measures the relative speed-up ratio compared to the base Lumina-mGPT model. 
For generation quality, we adopt \textbf{FID} to assess perceptual fidelity and diversity, and \textbf{CLIP-Score} to evaluate text–image semantic alignment performance. 

\begin{figure*}[t]
  \centering
  \includegraphics[width=0.975\linewidth, trim=2 2 2 2, clip]{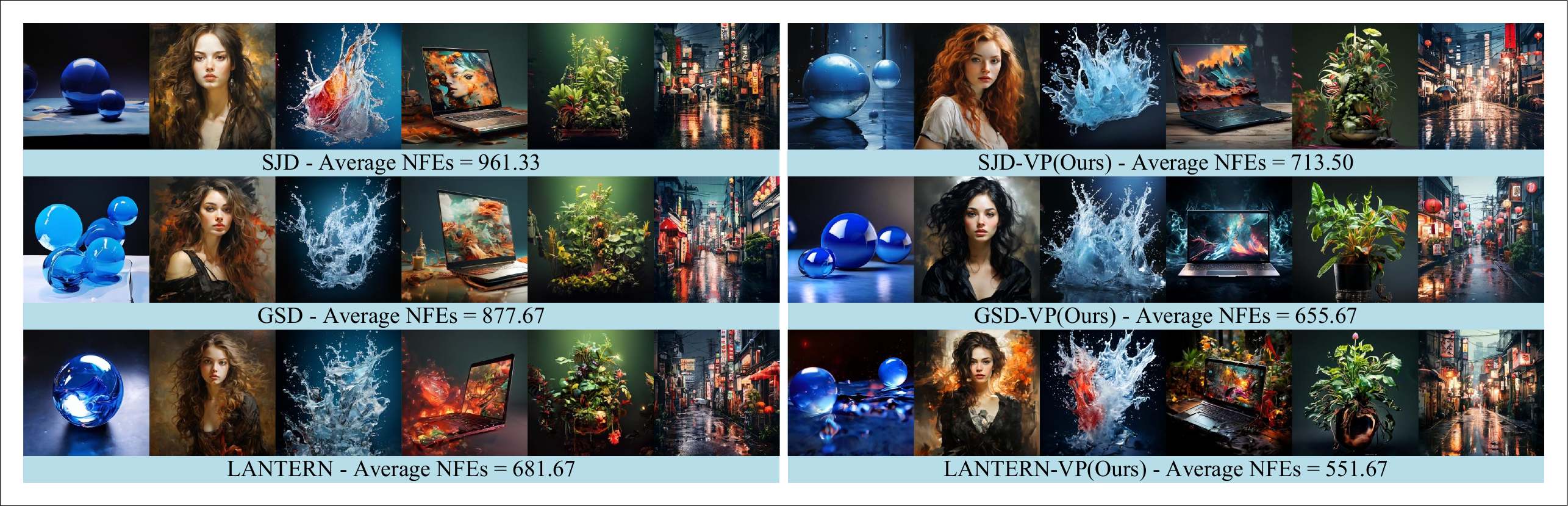}
  \vspace{-2mm}
  \caption{Qualitative experiment. Our method shows on average 1.3x NFE acceleration while maintaining image quality.}
  \label{fig:qual_all}
  \vspace{-1mm}
\end{figure*}

\paragraph{Implementation Details} 
All experiments are conducted on a single NVIDIA A100 GPU with 80 GB of memory. Our implementation is based on the official PyTorch and HuggingFace Transformers frameworks, with mixed-precision inference (FP16) enabled for all methods to ensure computational efficiency and fairness. Unless otherwise specified, we adopt the same hyperparameter configuration as GSD.
Specifically, the temperature is set to \(1.0\), and top-\(k\) sampling is applied with 2000. A classifier-free guidance scale of \(3.0\) is used to balance fidelity and diversity,
and the maximum generation length is fixed at \(8192\).
For SJD-VP, the historical memory length L= 3, and decay factor \(\gamma=0.8\).

\begin{table*}[t]
    \centering
    \small
    \renewcommand{\arraystretch}{0.8}
    \setlength{\tabcolsep}{0pt}
    \setlength{\abovecaptionskip}{2pt}
    \setlength{\belowcaptionskip}{2pt}
    \vspace{0pt}

    \begin{minipage}[t]{0.49\linewidth}
        \centering
        
        \captionof{table}{Ablation of Bayesian Fusion on PartPrompt.} 
        \label{tab:ablation_bayesian}
        \begin{tabular*}{\linewidth}{@{\hspace{4pt}\extracolsep{\fill}}>{\columncolor{white}}cccccc@{\hspace{4pt}}}
            \toprule
            \multicolumn{3}{c}{Sample Probability} & \multirow{2}{*}{\cellcolor{white}Latency} & \multirow{2}{*}{NFE} & \multirow{2}{*}{CLIP} \\
            $p_t$ & \hspace{15pt}$S_t$ &$p'_t$ &  &  &  \\
            \midrule
             \checkmark  & & & 36.07 s & 1035.3 & 32.09 \\
             & \hspace{15pt}\checkmark  & & 32.13 s & 899.7 & 32.10 \\
            & &  \checkmark & \textbf{27.00 s} & \textbf{720.7} & \textbf{32.11} \\
            \bottomrule
        \end{tabular*}

        \vspace{10pt}

        \captionof{table}{Ablation of Mask Growth Steps on PartPrompt.}
        \label{tab:ablation_mask}
        \begin{tabular*}{\linewidth}{@{\hspace{4pt}\extracolsep{\fill}}cccc@{\hspace{4pt}}}
            \toprule
            Growth Steps ($N$) & Latency & NFE & CLIP \\
            \midrule
            $N$=0 & 30.35 s & 831.2 & 32.09 \\
            $N$=1 & 28.21 s & 763.9 & 32.11 \\
            $N$=2 & 27.67 s & 744.1 & 32.12 \\
            \textbf{$N$=3} & \textbf{27.00 s} & \textbf{720.7} & \textbf{32.15} \\
            $N$=4 & 28.83 s & 785.5 & 32.10 \\
            \bottomrule
        \end{tabular*}

    \end{minipage}
    \hfill
    \begin{minipage}[t]{0.49\linewidth}
        \centering
        
        \captionof{table}{Evaluation across Various Model on PartPrompt.}
        \label{tab:ablation_model}
        \begin{tabular*}{\linewidth}{@{\hspace{4pt}\extracolsep{\fill}}cccc@{\hspace{4pt}}}
            \toprule
            Method & Latency & NFE & CLIP \\
            \midrule
            SJD (LlamaGen) & 19.86 s & 569.0 & 28.16 \\
            SJD-VP (LlamaGen) & \textbf{19.26 s} & \textbf{455. 39} & \textbf{28.18} \\
            \midrule
            SJD (Emu3) & 191.78 s & 3571.1 & 32.13 \\
            SJD-VP (Emu3) & \textbf{169.49 s} & \textbf{2511.19} & \textbf{32.14} \\
            \bottomrule
        \end{tabular*}
        
        \vspace{10pt}

        \captionof{table}{Ablation of Jacobi window sizes $W$ on PartPrompt.}
        \label{tab:ablation_W}
        \begin{tabular*}{\linewidth}{@{\hspace{4pt}\extracolsep{\fill}}cccc@{\hspace{4pt}}}
            \toprule
            Method & Latency & NFE & CLIP \\
            \midrule
            SJD ($W=32$) & 36.01 s & 1032.4 & 32.08 \\
            SJD-VP ($W=32$) & \textbf{21.68 s} & \textbf{537.6} & \textbf{32.08} \\
            \midrule
            SJD ($W=64$) & 36.07 s& 1035.2 & 32.10 \\
            SJD-VP ($W=64$) & \textbf{18.71 s} &  \textbf{401.7} & \textbf{32.11} \\
            \bottomrule
        \end{tabular*}
    \end{minipage}
    \vspace{-15pt}
\end{table*}


\subsection{Experimental Results}
\label{exp_result}
\paragraph{Quantitative Evaluation}
In Table~\ref{tab:main}, we present quantitative comparisons of our SJD-VP against state-of-the-art speculative decoding baselines on the Parti-prompt and MS-COCO 2017 benchmarks. It can be explicitly observed that SJD-VP, along with its variants, consistently outperform existing baselines. These results demonstrate a superior capability to drastically reduce decoding costs while preserving or even enhancing the visual and semantic fidelity. This suggests that our SJD-VP is effective in improving the efficiency–quality trade-off for autoregressive image generation and is plug-and-play, allowing it to be seamlessly integrated into existing speculative decoding methods and consistently delivering significant performance improvements.


\paragraph{Qualitative Evaluation}
To evaluate different aspects of generation performance, in Figure~\ref{fig:qual_all}, we present images generated by our method and baseline methods, conditioned on text prompts across five diverse categories: 
(1) \textbf{Counting}, assessing numerical accuracy and object fidelity; 
(2) \textbf{Portrait realism}, evaluating facial details and spatial consistency; 
(3) \textbf{Fast motion}, focusing on edge sharpness and temporal coherence; 
(4) \textbf{Indoor multi-object layout}, examining spatial relationships, occlusion handling, and material realism; and 
(5) \textbf{Outdoor night street}, assessing reflections, lighting effects, and overall scene coherence. Please refer to the Appendix~\ref{exp_setting} for the specific text prompts.
From Figure~\ref{fig:qual_all}, we can see that our methods consistently produce sharper details and fewer artifacts compared to baselines. Crucially, this improvement is achieved with an average \textbf{1.3$\times$ NFE reduction}, confirming that our SJD-VP substantially enhances both decoding efficiency and final image quality.

\subsection{Ablation Study}
\vspace{-0.55cm}

\noindent \paragraph{Does Bayesian Fusion yield superior performance?}
To answer this question, we sample draft tokens from different distributions $p_t$, $S_t$, and $p'_t$, which correspond to the original SJD, the prediction score, and the Bayesian Fusion, respectively. As shown in Table \ref{tab:ablation_bayesian}, the conditional prior ($S_t$) surpasses the baseline SJD ($p_t$), demonstrating that the  prior effectively captures valuable growth patterns. However, it still falls short of the Bayesian Fusion ($p'_t$). This indicates that while $S_t$ provides accurate structural guidance, it lacks the essential semantic context. Therefore, Bayesian Fusion is necessary to align these growth trends with the semantically valid distribution of $p_t$, ensuring optimal performance.
\noindent \paragraph{How does the number of consecutive growth steps $N$ impact performance?} To answer this, we analyze the sensitivity of the mask which forces SJD-VP to focus on tokens with an $N$-step consecutive growth. In Table~\ref{tab:ablation_mask}. The results exhibit an initial performance improvement followed by a decline. Within an optimal range (e.g., $N \le 3$), increasing $N$ enhances the reliability of the prior, as the multi-step constraint effectively filters out tokens that are not in a state of consecutive probability growth. However, further tightening the constraint (e.g., $N=4$) renders it overly strict. This leads to a scarcity of tokens that satisfy the continuous growth condition, thereby limiting the effective scope of the prior guidance and causing a regression in efficiency.
\vspace{-0.5cm}
\noindent \paragraph{How does SJD-VP perform across various models?} 
To evaluate our approach across different architectures, we compare SJD-VP against baseline SJD on two foundation models: LlamaGen~\cite{sun2024autoregressive} and Emu3~\cite{wang2024emu3}. As shown in Table~\ref{tab:ablation_model}, SJD-VP consistently achieves substantial speedups on both models. On the heavier Emu3 model, improvements are more pronounced due to larger computational footprint. Critically, generation quality remains stable across all models, with CLIP scores virtually unchanged. This shows our sampling strategy is broadly applicable and effectively accelerates inference.

\noindent \paragraph{How does the Jacobi window size $W$ affect performance?} 
To investigate the influence of the Jacobi window size $W$ on the efficiency of SJD-VP, we conduct ablations with $W=32$ and $W=64$. As shown in Table~\ref{tab:ablation_W}, both window configurations benefit from our SJD-VP. Notably, larger window sizes yield superior performance, as the expanded sampling range enables more effective verification prior. When $W=64$, SJD-VP achieves the most dramatic improvements in both latency and NFE while maintaining comparable generation quality. This phenomenon demonstrates that increased window sizes pair exceptionally well with our approach, delivering optimal efficiency and generation quality.

\section{Conclusions}

In this paper, we propose a novel Speculative Jacobi Decoding framework with verification prediction that enhances both the efficiency and quality of autoregressive image generation by aligning the drafting step with the verification step within the SJD decoding process. 
In particular, our method is plug-and-play, which can be seamlessly integrated into existing SJD variants, by simply replacing their draft steps while keeping all other components unchanged.   
Results show that our method achieves a superior efficiency–quality trade-off compared to state-of-the-art approaches.

\section*{Impact Statement}
This paper presents work whose goal is to advance the field of Machine Learning. There are many potential societal consequences of our work, none which we feel must be specifically highlighted here
\nocite{langley00}

\bibliography{example_paper}

@String(CVPR= {IEEE Conf. Comput. Vis. Pattern Recog.})

@String(CVPR  = {CVPR})

@article{brown2020language,
  title={Language models are few-shot learners},
  author={Brown, Tom and Mann, Benjamin and Ryder, Nick and Subbiah, Melanie and Kaplan, Jared D and Dhariwal, Prafulla and Neelakantan, Arvind and Shyam, Pranav and Sastry, Girish and Askell, Amanda and others},
  journal={Advances in neural information processing systems},
  volume={33},
  pages={1877--1901},
  year={2020}
}

@article{achiam2023gpt,
  title={Gpt-4 technical report},
  author={Achiam, Josh and Adler, Steven and Agarwal, Sandhini and Ahmad, Lama and Akkaya, Ilge and Aleman, Florencia Leoni and Almeida, Diogo and Altenschmidt, Janko and Altman, Sam and Anadkat, Shyamal and others},
  journal={arXiv preprint arXiv:2303.08774},
  year={2023}
}

@article{touvron2023llama,
  title={Llama: Open and efficient foundation language models},
  author={Touvron, Hugo and Lavril, Thibaut and Izacard, Gautier and Martinet, Xavier and Lachaux, Marie-Anne and Lacroix, Timoth{\'e}e and Rozi{\`e}re, Baptiste and Goyal, Naman and Hambro, Eric and Azhar, Faisal and others},
  journal={arXiv preprint arXiv:2302.13971},
  year={2023}
}

@inproceedings{chen2020generative,
  title={Generative pretraining from pixels},
  author={Chen, Mark and Radford, Alec and Child, Rewon and Wu, Jeffrey and Jun, Heewoo and Luan, David and Sutskever, Ilya},
  booktitle={International conference on machine learning},
  pages={1691--1703},
  year={2020},
  organization={PMLR}
}

@article{ramesh2022hierarchical,
  title={Hierarchical text-conditional image generation with clip latents},
  author={Ramesh, Aditya and Dhariwal, Prafulla and Nichol, Alex and Chu, Casey and Chen, Mark},
  journal={arXiv preprint arXiv:2204.06125},
  volume={1},
  number={2},
  pages={3},
  year={2022}
}

@article{van2016conditional,
  title={Conditional image generation with pixelcnn decoders},
  author={Van den Oord, Aaron and Kalchbrenner, Nal and Espeholt, Lasse and Vinyals, Oriol and Graves, Alex and others},
  journal={Advances in neural information processing systems},
  volume={29},
  year={2016}
}

@inproceedings{leviathan2023fast,
  title={Fast inference from transformers via speculative decoding},
  author={Leviathan, Yaniv and Kalman, Matan and Matias, Yossi},
  booktitle={International Conference on Machine Learning},
  pages={19274--19286},
  year={2023},
  organization={PMLR}
}

@article{sun2024autoregressive,
  title={Autoregressive model beats diffusion: Llama for scalable image generation},
  author={Sun, Peize and Jiang, Yi and Chen, Shoufa and Zhang, Shilong and Peng, Bingyue and Luo, Ping and Yuan, Zehuan},
  journal={arXiv preprint arXiv:2406.06525},
  year={2024}
}

@article{wang2024emu3,
  title={Emu3: Next-token prediction is all you need},
  author={Wang, Xinlong and Zhang, Xiaosong and Luo, Zhengxiong and Sun, Quan and Cui, Yufeng and Wang, Jinsheng and Zhang, Fan and Wang, Yueze and Li, Zhen and Yu, Qiying and others},
  journal={arXiv preprint arXiv:2409.18869},
  year={2024}
}

@inproceedings{van2016pixel,
  title={Pixel recurrent neural networks},
  author={Van Den Oord, A{\"a}ron and Kalchbrenner, Nal and Kavukcuoglu, Koray},
  booktitle={International conference on machine learning},
  pages={1747--1756},
  year={2016},
  organization={PMLR}
}

@inproceedings{esser2021taming,
  title={Taming transformers for high-resolution image synthesis},
  author={Esser, Patrick and Rombach, Robin and Ommer, Bjorn},
  booktitle={Proceedings of the IEEE/CVF conference on computer vision and pattern recognition},
  pages={12873--12883},
  year={2021}
}

@inproceedings{ramesh2021zero,
  title={Zero-shot text-to-image generation},
  author={Ramesh, Aditya and Pavlov, Mikhail and Goh, Gabriel and Gray, Scott and Voss, Chelsea and Radford, Alec and Chen, Mark and Sutskever, Ilya},
  booktitle={International conference on machine learning},
  pages={8821--8831},
  year={2021},
  organization={Pmlr}
}

@inproceedings{yu2024magvit2,
  title={Magvit-2: Masked generative video transformer},
  author={Yu, Jiahui and Chang, Huiwen and Milan, Anton and et al.},
  booktitle={CVPR},
  year={2024}
}

@inproceedings{wang2025parallelized,
  title={Parallelized autoregressive visual generation},
  author={Wang, Yuqing and Ren, Shuhuai and Lin, Zhijie and Han, Yujin and Guo, Haoyuan and Yang, Zhenheng and Zou, Difan and Feng, Jiashi and Liu, Xihui},
  booktitle={Proceedings of the Computer Vision and Pattern Recognition Conference},
  pages={12955--12965},
  year={2025}
}

@article{jang2024lantern,
  title={Lantern: Accelerating visual autoregressive models with relaxed speculative decoding},
  author={Jang, Doohyuk and Park, Sihwan and Yang, June Yong and Jung, Yeonsung and Yun, Jihun and Kundu, Souvik and Kim, Sung-Yub and Yang, Eunho},
  journal={arXiv preprint arXiv:2410.03355},
  year={2024}
}

@article{liu2024lumina,
  title={Lumina-mgpt: Illuminate flexible photorealistic text-to-image generation with multimodal generative pretraining},
  author={Liu, Dongyang and Zhao, Shitian and Zhuo, Le and Lin, Weifeng and Xin, Yi and Li, Xinyue and Qin, Qi and Qiao, Yu and Li, Hongsheng and Gao, Peng},
  journal={arXiv preprint arXiv:2408.02657},
  year={2024}
}

@inproceedings{wu2025janus,
  title={Janus: Decoupling visual encoding for unified multimodal understanding and generation},
  author={Wu, Chengyue and Chen, Xiaokang and Wu, Zhiyu and Ma, Yiyang and Liu, Xingchao and Pan, Zizheng and Liu, Wen and Xie, Zhenda and Yu, Xingkai and Ruan, Chong and others},
  booktitle={Proceedings of the Computer Vision and Pattern Recognition Conference},
  pages={12966--12977},
  year={2025}
}

@article{ren2025beyond,
  title={Beyond next-token: Next-x prediction for autoregressive visual generation},
  author={Ren, Sucheng and Yu, Qihang and He, Ju and Shen, Xiaohui and Yuille, Alan and Chen, Liang-Chieh},
  journal={arXiv preprint arXiv:2502.20388},
  year={2025}
}

@article{llamagen,
  title={Autoregressive model beats diffusion: Llama for scalable image generation},
  author={Sun, Peize and Jiang, Yi and Chen, Shoufa and Zhang, Shilong and Peng, Bingyue and Luo, Ping and Yuan, Zehuan},
  journal={arXiv preprint arXiv:2406.06525},
  year={2024}
}

@article{luminangpt,
  title={Lumina-mgpt: Illuminate flexible photorealistic text-to-image generation with multimodal generative pretraining},
  author={Liu, Dongyang and Zhao, Shitian and Zhuo, Le and Lin, Weifeng and Qiao, Yu and Li, Hongsheng and Gao, Peng},
  journal={arXiv preprint arXiv:2408.02657},
  year={2024}
}

@article{SJD,
  title={Accelerating auto-regressive text-to-image generation with training-free speculative jacobi decoding},
  author={Teng, Yao and Shi, Han and Liu, Xian and Ning, Xuefei and Dai, Guohao and Wang, Yu and Li, Zhenguo and Liu, Xihui},
  journal={arXiv preprint arXiv:2410.01699},
  year={2024}
}

@article{LANTERN,
  title={Lantern: Accelerating visual autoregressive models with relaxed speculative decoding},
  author={Jang, Doohyuk and Park, Sihwan and Yang, June Yong and Jung, Yeonsung and Yun, Jihun and Kundu, Souvik and Kim, Sung-Yub and Yang, Eunho},
  journal={arXiv preprint arXiv:2410.03355},
  year={2024}
}

@article{so2025mc,
  title={MC-SJD: Maximal Coupling Speculative Jacobi Decoding for Autoregressive Visual Generation Acceleration},
  author={So, Junhyuk and Kook, Hyunho and Jang, Chaeyeon and Park, Eunhyeok},
  journal={arXiv preprint arXiv:2510.24211},
  year={2025}
}

@inproceedings{jacobi,
  title={Accelerating feedforward computation via parallel nonlinear equation solving},
  author={Song, Yang and Meng, Chenlin and Liao, Renjie and Ermon, Stefano},
  booktitle={International Conference on Machine Learning},
  pages={9791--9800},
  year={2021},
  organization={PMLR}
}

@inproceedings{mscoco,
  title={Microsoft coco: Common objects in context},
  author={Lin, Tsung-Yi and Maire, Michael and Belongie, Serge and Hays, James and Perona, Pietro and Ramanan, Deva and Doll{\'a}r, Piotr and Zitnick, C Lawrence},
  booktitle={Computer vision--ECCV 2014: 13th European conference, zurich, Switzerland, September 6-12, 2014, proceedings, part v 13},
  pages={740--755},
  year={2014},
  organization={Springer}
}

@inproceedings{clip,
  title={Learning transferable visual models from natural language supervision},
  author={Radford, Alec and Kim, Jong Wook and Hallacy, Chris and Ramesh, Aditya and Goh, Gabriel and Agarwal, Sandhini and Sastry, Girish and Askell, Amanda and Mishkin, Pamela and Clark, Jack and others},
  booktitle={International conference on machine learning},
  pages={8748--8763},
  year={2021},
  organization={PmLR}
}

@article{FID,
  title={Gans trained by a two time-scale update rule converge to a local nash equilibrium},
  author={Heusel, Martin and Ramsauer, Hubert and Unterthiner, Thomas and Nessler, Bernhard and Hochreiter, Sepp},
  journal={Advances in neural information processing systems},
  volume={30},
  year={2017}
}

@inproceedings{so2025grouped,
  title={Grouped Speculative Decoding for Autoregressive Image Generation},
  author={So, Junhyuk and Shin, Juncheol and Kook, Hyunho and Park, Eunhyeok},
  booktitle={Proceedings of the IEEE/CVF International Conference on Computer Vision},
  pages={15375--15384},
  year={2025}
}

@misc{liu2024lumina-mgpt,
      title={Lumina-mGPT: Illuminate Flexible Photorealistic Text-to-Image Generation with Multimodal Generative Pretraining},
      author={Dongyang Liu and Shitian Zhao and Le Zhuo and Weifeng Lin and Yu Qiao and Hongsheng Li and Peng Gao},
      year={2024},
      eprint={2408.02657},
      archivePrefix={arXiv},
      primaryClass={cs.CV},
      url={https://arxiv.org/abs/2408.02657},
}

@inproceedings{van2017neural,
  title={Neural discrete representation learning},
  author={van den Oord, Aaron and Vinyals, Oriol and Kavukcuoglu, Koray},
  booktitle={Advances in Neural Information Processing Systems},
  year={2017}
}

@inproceedings{razavi2019generating,
  title={Generating diverse high-fidelity images with vq-vae-2},
  author={Razavi, Ali and van den Oord, Aaron and Vinyals, Oriol},
  booktitle={Advances in Neural Information Processing Systems},
  year={2019}
}

@inproceedings{chang2022maskgit,
  title={Maskgit: Masked generative image transformer},
  author={Chang, Huiwen and Zhang, Han and Barber, Jarred and Maschinot, Aaron and Lezama, Jose and Jiang, Lu and Yang, Ming-Hsuan and Murphy, Kevin and Freeman, William T and Rubinstein, Michael and others},
  booktitle={Proceedings of the IEEE/CVF Conference on Computer Vision and Pattern Recognition},
  year={2022}
}

@article{yu2022scaling,
  title={Scaling autoregressive models for content-rich text-to-image generation},
  author={Yu, Jiahui and Xu, Yuanzhong and Koh, Jing Yu and Luong, Thang and Baid, Gunjan and Wang, Zirui and Vasudevan, Vijay and Ku, Alexander and Yang, Yinfei and Ayan, Burcu Karagol and others},
  journal={arXiv preprint arXiv:2206.10789},
  year={2022}
}

@inproceedings{yu2018generative,
  title={Generative image inpainting with contextual attention},
  author={Yu, Jiahui and Lin, Zhe and Yang, Jimei and Shen, Xiaohui and Lu, Xin and Huang, Thomas S},
  booktitle={Proceedings of the IEEE conference on computer vision and pattern recognition},
  year={2018}
}

@inproceedings{alayrac2022flamingo,
  title={Flamingo: a visual language model for few-shot learning},
  author={Alayrac, Jean-Baptiste and Donahue, Jeff and Luc, Pauline and Miech, Antoine and Barr, Iain and Hasson, Yana and Lenc, Karel and Mensch, Arthur and Millican, Katherine and Reynolds, Malcolm and others},
  booktitle={Advances in Neural Information Processing Systems},
  year={2022}
}

@inproceedings{liu2024visual,
  title={Visual instruction tuning},
  author={Liu, Haotian and Li, Chunyuan and Wu, Qingyang and Lee, Yong Jae},
  booktitle={Advances in neural information processing systems},
  year={2024}
}

@article{team2023gemini,
  title={Gemini: a family of highly capable multimodal models},
  author={Team, Gemini and Anil, Rohan and Borgeaud, Sebastian and Wu, Yonghui and Alayrac, Jean-Baptiste and Yu, Jiahui and Soricut, Radu and Schalkwyk, Johan and Dai, Andrew M and Hauth, Anja and others},
  journal={arXiv preprint arXiv:2312.11805},
  year={2023}
}
\bibliographystyle{icml2026}

\newpage
\onecolumn
\appendix
\section{Detail Experiment Settings and Results Interpretation}
\label{exp_setting}
\noindent \paragraph{Prompt of qualitative experiment}
All qualitative results in both the main paper and this supplementary material are generated using the following prompts:

\begin{itemize}
    \item \textbf{Counting / numeracy:} \textit{2 blue spheres}
    \item \textbf{Portrait realism:} \textit{female portrait}
    \item \textbf{Fast motion:} \textit{splash frozen}
    \item \textbf{Indoor object layout:} \textit{laptop}, \textit{plant}
    \item \textbf{Outdoor night street:} \textit{rainy Tokyo street}
\end{itemize}

\noindent \paragraph{Results Interpretation of GSD Acceleration}  
We observe in Table~\ref{tab:main} that SJD-VP exhibits a discrepancy between NFE and latency improvements when accelerating GSD. Specifically, while NFE shows noticeable reduction, the corresponding latency decrease is less pronounced. Through detailed analysis, we identify that this discrepancy stems from GSD's inherent sorting overhead during verification, which significantly increases the per-NFE execution time. In contrast, other baseline methods do not incur such sorting costs during their verification phases. Therefore, the mismatch between NFE and latency reduction observed in GSD is a well-founded phenomenon that reflects the architectural characteristics of the method.

\section{Supplementary Theoretical Derivations}
\label{sec:appendix_proof}

In this section, we provide the detailed mathematical steps omitted in Section 4.4, specifically for Eq. (\ref{taylor}) (Taylor Expansion) and Eq.(\ref{eq14}) (Expectation Decomposition) and further elaborate on the proof conclusion in Eq.(\ref{eq16}).

\subsection{Derivation of Eq.~(\ref{taylor}) (Taylor Expansion)}
In section 4.4, we perform a first-order Taylor expansion at $m=0$ and further simplify the $D_{\text{TV}}(p'_t, q)$ as
\begin{equation}
\small
    D_{\text{TV}}(p'_t, q) = \frac{1}{2} \sum_x |p_t(x) + m \cdot \omega(x) \hat{y}(x) - q(x)| \approx D_{\text{TV}}(p_t, q) + m \frac{d}{dm} D_{\text{TV}}(p'_t, q) \Big|_{m=0}
\end{equation}
Specifically, the derivative term $\frac{d}{dm} D_{\text{TV}}(p'_t, q)$ represents the rate of change of the distance with respect to the perturbation magnitude. Here, to facilitate the derivation, the term $ p_t(x) - q(x) + m \cdot \omega(x)\hat{y}(x)$ is defined as $u(x)$. By applying the chain rule and the derivative formula (i.e., $\frac{\partial |u|}{\partial u} = \text{sign}(u)$), we have:
\begin{equation}
\begin{split}
    \frac{d}{dm} D_{\text{TV}}(p'_t, q) &= \frac{1}{2} \sum_{x} \frac{d}{dm} \left| u(x) \right| \\
    &= \frac{1}{2} \sum_{x} \frac{\partial |u(x)|}{\partial u(x)} \cdot \frac{\partial u(x)}{\partial m} \\
    &= \frac{1}{2} \sum_{x} \text{sign}(u(x)) \cdot \omega(x)\hat{y}(x),
\end{split}
\end{equation}
Substituting the definition of $u(x) = p_t(x) - q(x) + m \cdot \omega(x)\hat{y}(x)$, we obtain:
\begin{equation}
    \frac{d}{dm} D_{\text{TV}}(p'_t, q) = \frac{1}{2} \sum_{x} \text{sign}\big(p_t(x) - q(x) + m \cdot \omega(x)\hat{y}(x)\big) \cdot \omega(x)\hat{y}(x).
\end{equation}
Finally, by setting $m=0$, we obtain the result presented in the main text:
\begin{equation}
    D_{\text{TV}}(p'_t, q) \approx D_{\text{TV}}(p_t, q) + \frac{m}{2} \sum_x \text{sign}(p_t(x)-q(x)) \cdot \omega(x) \hat{y}(x).
\end{equation}

\subsection{Derivation of Eq.~(\ref{eq14}) (Expectation Decomposition) and Eq.~(\ref{eq16}) (Proof Conclusion)}
Next, we derive the Expectation Decomposition in Eq.~(\ref{eq14}). By applying the standard expectation decomposition formula $\mathbb{E}[AB] = \mathbb{E}[A]\mathbb{E}[B] + \text{Cov}(A, B)$, we expand the interaction term between  $y(x)\hat{y}(x)$ and  $\omega(x)$ as follows:
\begin{equation}
\begin{split}
    D_{\text{TV}}(p'_t, q) &\approx D_{\text{TV}}(p_t, q) - \frac{m}{2} \mathbb{E}_{x \sim p_t}[y(x)\hat{y}(x)\omega(x)] \\
    &= D_{\text{TV}}(p_t, q) - \frac{m}{2} \Big( \mathbb{E}_{x \sim p_t}[y(x)\hat{y}(x)] \cdot \mathbb{E}_{x \sim p_t}[\omega(x)] + \text{Cov}\big(y(x)\hat{y}(x), \omega(x)\big) \Big).
\end{split}
\end{equation}
Finally, we justify the approximation $\text{Cov}(y(x)\hat{y}(x), \omega(x)) \approx 0$ in Eq.(\ref{eq16}). The reason is that our weighting mechanism is designed to be \textbf{agnostic to the prediction result (i.e., $y(x)\hat{y}(x)$)}, relying solely on probability growth trajectory. While this strategy achieves high precision,  the magnitude of the assigned weights $\omega(x)$ is driven by the pattern of probability growth rather than verification results. This structural independence allows us to treat the weight generation and the alignment term as uncorrelated variables, rendering the covariance negligible (i.e., $\text{Cov}(y(x)\hat{y}(x), \omega(x)) \approx 0$).

\section{Supplementary Ablation Study}
In this section, we discuss the performance of SJD-VP across different Jacobi window sizes $W$ and various model architectures. Furthermore, we provide an analysis of the additional computational overhead introduced by the SJD-VP.

\noindent \paragraph{How does the EWA Decay Factor $\gamma$ impact Verification Prediction?}
 To answer this question, we evaluate EWA Decay Factor $\gamma$ across a range of values. The results shown in Table~\ref{tab:ablation_decay} indicate that performance degrades at both extremes. A lower $\gamma$ (e.g., 0.2, 0.4) results in rapid memory decay, making the predictor overly sensitive to recent iterations. Conversely, a higher $\gamma$ (e.g., 1.0) introduces excessive lag, causing the prediction to drift behind the actual trend due to high inertia. The optimal balance is achieved at $\gamma = 0.8$, where the predictor retains sufficient history while remaining responsive to new growth patterns.

\begin{table*}[h]
    \centering
    \small
    \renewcommand{\arraystretch}{0.8}
    \setlength{\tabcolsep}{12pt}
    \setlength{\abovecaptionskip}{2pt}
    \setlength{\belowcaptionskip}{2pt}
    \caption{Ablation of Decay Factor $\gamma$ on PartPrompt.}
    \label{tab:ablation_decay}
    \begin{tabular}{cccc}
        \toprule
        $\gamma$ & Latency & NFE & CLIP \\
        \midrule
        1.0 & 28.75 s & 781.6 & 32.12 \\
        \textbf{0.8} & \textbf{27.00 s} & \textbf{720.7} & \textbf{32.15} \\
        0.6 & 27.51 s & 737.8 & 32.11 \\
        0.4 & 27.92 s & 752.0 & 32.09 \\
        0.2 & 28.04 s & 756.9 & 32.12 \\
        \bottomrule
    \end{tabular}
\end{table*}

\noindent \paragraph{How does the Top-$k$ filtering ratio impact performance?} To answer this, we investigate the influence of the Top-$k$ filtering ratio on performance of SJD-VP.
 As shown in Table~\ref{tab:ablation_topk}, stricter filtering yields a dual benefit. First, it significantly reduces redundant computational overhead by strategically limiting the sampling scope to only the most promising candidates, directly leading to lower latency. Second, it enhances robustness by aggressively pruning low-probability tail noise, preventing potential interference from irrelevant tokens. Consequently, the setting of $k=10\%$ achieves the optimal balance, delivering the fastest inference speed without compromising generation quality.

 \begin{table*}[h]
    \centering
    \small
    \renewcommand{\arraystretch}{0.8}
    \setlength{\tabcolsep}{10pt}
    \setlength{\abovecaptionskip}{2pt}
    \setlength{\belowcaptionskip}{2pt}
    \caption{Ablation of the Top-$k$ filtering strategy on PartPrompt.}
    \label{tab:ablation_topk}
    \begin{tabular}{cccc}
        \toprule
        Filter Top-$k$ & Latency & NFE & CLIP \\
        \midrule
        w/o Filter & 32.19 s & 904.8 & 32.09 \\
        $k=70\%$ & 30.02 s & 833.6 & 32.10 \\
        $k=50\%$ & 28.47 s & 759.0 & 33.10 \\
        $k=30\%$ & 27.51 s & 739.4 & 32.11 \\
        \textbf{$k=10\%$} & \textbf{27.00 s} & \textbf{720.7} & \textbf{32.15} \\
        \bottomrule
    \end{tabular}
\end{table*}

\noindent \paragraph{What is the computational overhead of SJD-VP?} To assess the specific cost of our proposed modules, we measure the additional computational burden imposed by different methods. As reported in Table~\ref{tab:computational_analysis}, SJD-VP incurs a minimal per-step overhead of 1.026 ms, confirming that our Bayesian Fusion and Prior Estimation modules are lightweight.

Furthermore, we analyze the behavior of GSD. While GSD effectively reduces NFE, it introduces a significantly higher overhead of 1.967 ms per step—nearly double that of SJD-VP. This is likely attributed to its internal reliance on logits sorting and re-ranking operations, which are computationally more demanding. This finding clarifies the discrepancy between NFE reduction and actual speedup. This observation also provides empirical support for the analysis presented in Section~\ref{exp_setting}.

\begin{table*}[h]
    \centering
    \small
    \setlength{\tabcolsep}{8pt}
    \renewcommand{\arraystretch}{0.8}
    \setlength{\abovecaptionskip}{2pt}
    \setlength{\belowcaptionskip}{2pt}
    \caption{Computational Analysis.}
    \label{tab:comprehensive_results}
    \begin{tabular}{lccc}
        \toprule
         & SJD-VP & GSD & GSD-VP \\
        \midrule
        Per-step overhead & 1.026ms & 1.967 ms & 2.926 ms\\
        \bottomrule
    \end{tabular}
\end{table*}


\end{document}